# Learning Through Dialogue: Engagement and Efficacy Matter More Than Explanations

**Shaz Furniturewala[1], Gerard Christopher Yeo[1,2], Kokil Jaidka[1,3]**
[1] Centre for Trusted Internet and Community,
National University of Singapore, Singapore
[2]Agency for Science, Technology and Research, Singapore
[3]Department of Communications and New Media,
National University of Singapore, Singapore
**Correspondence:** jaidka@nus.edu.sg

## Abstract

Large language models (LLMs) are increasingly used as conversational partners for learning, yet the interactional dynamics supporting users' learning and engagement are understudied. We analyze the linguistic and interactional features from both LLM and participant chats across 397 human–LLM conversations about socio-political issues to identify the mechanisms and conditions under which LLM explanations shape changes in political knowledge and confidence. Mediation analyses reveal that LLM explanatory richness partially supports confidence by fostering users' reflective insight, whereas its effect on knowledge gain operates entirely through users' cognitive engagement. Moderation analyses show that these effects are highly conditional and vary by political efficacy. Confidence gains depend on how high-efficacy users experience and resolve uncertainty. Knowledge gains depend on high-efficacy users' ability to leverage extended interaction, with longer conversations benefiting primarily reflective users. In summary, we find that learning from LLMs is an interactional achievement, not a uniform outcome of better explanations. The findings underscore the importance of aligning LLM explanatory behavior with users' engagement states to support effective learning in designing Human-AI interactive systems.

## 1 Introduction

Large language models (LLMs) are increasingly embedded in everyday political information-seeking (NiemanLab, 2024). This shift is especially salient in 2024, a year marked by political uncertainty and widespread regime change, with national elections taking place in 64 countries worldwide. As citizens navigate complex and contested political environments, information-seeking conversations play a central role in how political knowledge is formed, revised, and evaluated. Chatbots such as ChatGPT now mediate a growing share of these interactions, reshaping how millions of users search for political information and potentially influencing how people "think, read, and remember." (Marr, 2023) While early evidence suggests that users increasingly prefer chatbot-assisted search (Kan, 2023), **there remains a notable lack of publicly available corpora paired to user demographics and learning outcomes.**

Recent advances have transformed LLMs from static retrieval tools into conversational agents capable of sustained dialogue (Wang et al., 2024a). Beyond answering isolated questions, LLMs can provide explanations, respond to follow-up queries, and adapt their responses across turns in ways that resemble collaborative learning interactions (Skjuve et al., 2024). These capabilities position LLMs as active participants in political sensemaking rather than passive sources of information. Users approach these interactions with heterogeneous goals, ranging from brief factual clarification to exploratory reasoning and sensemaking, **further underscoring the need to examine interactional dynamics across conversations** rather than focusing solely on individual responses.

Political information-seeking is closely tied to political learning. Developing political understanding often requires synthesizing heterogeneous evidence, weighing competing claims, and reasoning under uncertainty (Bode, 2016). Conversational interaction with an LLM may support these processes by enabling users to ask clarifying questions, explore alternative explanations, and reflect on their understanding (Oppenheimer et al., 2025). Such interactions may facilitate both knowledge acquisition and calibration of confidence. Whether these benefits materialize, however, depends on **how explanations are structured and how users engage with them during dialogue.**

This gap is consequential because learning is not a unidirectional process, but one that unfolds through the interplay between explanations and learner engagement (Lukita et al., 2017; Xhemajli, 2016). Explanations vary in depth, structure, and linguistic form, shaping how information is processed and integrated. Learners actively influence learning through behaviors such as asking follow-up questions, expressing uncertainty, reflecting on explanations, and coordinating turn-taking. These engagement processes are especially important in political contexts, where understanding is often constructed through exploration rather than simple acceptance of information.

In this work, we adopt an interactional perspective on political learning from human–LLM conversations. We focus on two learning-related outcomes: changes in political knowledge and changes in confidence in one's understanding. Specifically, we ask the following research questions:

**RQ1**: Which linguistic and explanatory features of LLM responses are associated with changes in users' political knowledge and confidence?

**RQ2**: Do human engagement processes during conversation mediate the relationship between LLM explanations and learning outcomes?

**RQ3**: Do individual differences (e.g., political efficacy) and interactional cues (e.g., confusion, conversational depth, or interaction length) moderate the effects of LLM explanations on learning?

To address these questions, we analyze 397 political conversations between users and a chatbot. We computationally extract linguistic, explanatory, and interactional features from both LLM and human turns using established lexicons and automated text analysis models. We then employ multilevel regression models, complemented by mediation and moderation analyses, to disentangle how LLM explanatory characteristics and human engagement jointly—and conditionally—shape changes in political knowledge and confidence. This work makes two primary contributions:

- We show that political learning from LLMs depends not only on explanatory quality, but also on how users engage during interaction.
- We clarify both the processes (how learning occurs) and the conditions (when learning is most likely) underlying human–LLM political dialogue through integrated mediation and moderation analyses.

## 2 Related Work

### 2.1 LLMs for Learning and Educational Support

Large language models (LLMs) have been widely studied as tools for educational support due to their ability to generate fluent, context-sensitive, and domain-adaptive responses (Chu et al., 2025; Wang et al., 2024b). Systematic reviews indicate that LLM-based applications span multiple domains and are often associated with improvements in task performance and learner engagement (Dong et al., 2024) in areas such as mathematics, writing, and problem-solving (Yan et al., 2025).

More recent studies situate LLMs within intelligent tutoring systems or as personalized tutors, emphasizing adaptive feedback and scaffolding (Park et al., 2024; Kasneci et al., 2023). Despite this progress, much of the literature continues to conceptualize LLMs as one-way instructional tools rather than interactive partners.Our work addresses this gap by examining how explanatory properties of LLM responses interact with users' conversational engagement to shape learning outcomes in open-ended dialogue.

### 2.2 Learning Through Interactive Engagement

Educational research has long emphasized that learning is fundamentally interactive. Interactive pedagogies frame learning as a dialogic process involving reflection, elaboration, and sense-making, linking engagement behaviors to gains in critical thinking and metacognition (Divakova, 2011; Strebna and Sotsenko, 2007; Blyznyuk and Kachak, 2024). This perspective aligns with the *computers as social actors* (CASA) paradigm, which shows that people routinely apply social norms from human–human interaction to computational systems that display interactive or human-like cues (Nass et al., 1994). Recent work demonstrates that users treat such systems as conversational partners, attributing agency and responding with social behaviors that shape trust, engagement, and perceptions (Xu et al., 2022; Zhou et al., 2023; Song et al., 2022; Munnukka et al., 2022; Yu et al., 2024).

Despite these insights, engagement has rarely been operationalized computationally in studies of LLM-mediated learning. Existing work typically treats engagement indirectly (e.g., time-on-task) or overlooks it altogether. Consequently, little is known about whether engagement explains how LLM explanations influence learning (mediation) or when these explanations are most effective (moderation). We address this gap by extracting engagement-related features from dialogue and testing both their mediating and moderating roles.

### 2.3 Human–LLM Conversational Datasets

Recent years have seen rapid growth in datasets capturing natural human–LLM interaction. Large-scale corpora such as InfinityChat (Jiang et al.) and LMSYS-Chat-1M (Chiang et al., 2024), along with WildChat and ShareChat (Zhao et al.; Yan et al., 2025), provide millions of in-the-wild conversations across diverse contexts and models. These resources are invaluable for studying interaction patterns and model behavior.

However, they lack paired outcome measures (e.g., pre–post knowledge or confidence), fine-grained engagement signals, and experimental structure needed to examine learning processes and boundary conditions. As a result, they cannot directly address when or how LLM explanations support learning. Our study complements this work by analyzing conversations linked to learning outcomes and engagement features, enabling process-oriented analysis of learning in dialogue.

**In summary.** Prior research has established that LLMs offer interactive engagement that is central to learning, but these elements are rarely disambiguated. Existing datasets and evaluation frameworks do not support analyses of how explanatory features, engagement, and user characteristics jointly shape learning outcomes. These gaps have direct implications for analysis: if explanations operate through interactional processes rather than exerting uniform effects, approaches focused only on main effects are insufficient. Mediation helps clarify *how* explanatory features influence outcomes through engagement, while moderation helps identify *when* and *for whom* explanatory dialogue is effective.

To address these gaps, we introduce **Arthos**, a purpose-built experimental chatbot system designed to elicit extended political dialogue while enabling fine-grained measurement of explanation, engagement, and pre–post learning outcomes. This design allows us to examine learning as an interactional process rather than a property of model outputs alone. We describe the system architecture and experimental setup in detail in the following section.

## 3 Methodology

### 3.1 Chatbot System Description

**System Goals and Experimental Constraints.** We developed a chatbot with the primory goal being to elicit naturalistic human–LLM conversations that resemble real-world political information seeking, while allowing us to enforce the quality of the discussion through controlled experimental settings and the measurement of behavior, engagement dynamics, and learning outcomes. To achieve this, the system needed to (a) sustain multi-turn interaction, (b) integrate seamlessly with survey-based pre–post measurement, and (c) operate under strict institutional privacy and ethical constraints appropriate for politically sensitive data.

**Model Deployment and System Configuration.** We deployed Meta's LLaMA-3.1-70B (Versatile) model via Groq's inference-only API. This setup enabled fast, long-context inference suitable for extended dialogue while maintaining separation from commercial data storage or training pipelines. The system was configured with a context window of up to 128,000 tokens and a maximum output length of 32,000 tokens, allowing conversations to unfold without truncation. The temperature parameter was set to the default value of 1.0 to balance coherence with linguistic variability. All interactions were routed through a custom backend server that handled prompt construction, request routing, and secure logging.

**Prompt Structure and Tutoring Behavior.** The chatbot operated in a retrieval-augmented fashion. Each session was initialized with a structured prompt containing two components: (1) a curated set of news articles relevant to the participant's assigned experimental condition, and (2) a set of target questions previously completed in the baseline survey. The model was instructed to act as a tutoring assistant rather than an answer generator. It guided participants through evaluating evidence, comparing competing interpretations, and reasoning through the questions using the provided sources, without directly supplying final answers. This design ensured that learning outcomes reflected interactional reasoning processes rather than exposure to authoritative responses. The detailed prompts provided to the chatbots are reported in Appendix B.

### 3.2 Data Collection

**Study Design and Context.** We analyze conversational data collected as part of a pre–post political learning experiment. Participants engaged in multi-turn discussions with a chatbot about contemporary political issues, allowing us to observe how conversational dynamics relate to changes in political knowledge and confidence. The study was conducted in two national contexts: the United States and India. Conversations were conducted in English for all participants, with an additional Spanish condition for U.S. participants.

**Experimental Conditions and Conversation Setup.** Participants were randomly assigned to a country–issue–party condition. Issues were drawn from salient public policy domains, including crime rates, air quality, and unemployment, and paired with the policy position of a political party relevant to the participant's national context.

**Procedure and Outcome Measures.** At the start of the session, participants completed a *pre-conversation survey* measuring (a) issue-specific political knowledge and (b) confidence in their understanding of the issue. Participants then interacted freely with the chatbot, with encouragement to ask follow-up questions and probe points of confusion. Following the dialogue, participants completed a *post-conversation survey* that repeated the same knowledge and confidence measures. Exemplar questions are provided in Appendix C.

From these responses, we computed two primary outcomes: **knowledge gain**, defined as the difference between post- and pre-conversation knowledge scores, and **confidence change**, defined as the difference between post- and pre-conversation confidence ratings.

**Dataset Composition.** The final dataset consists of 397 conversations from 152 participants. Of these, 305 conversations concern U.S. politics (195 in English and 110 in Spanish), and 92 concern Indian politics (all in English). Conversation transcripts were processed to extract linguistic, explanatory, and interactional features used in subsequent analyses.

| **Participants** | 152 |
|---|---|
| **Conversations** | 397 |
| U.S. (English) | 195 |
| U.S. (Spanish) | 110 |
| India (English) | 92 |
| **Average Human Turns per Conversation** (4286 Total Turns, 52081 Total Words) | |
| U.S. (English) | 5.4 |
| U.S. (Spanish) | 4.7 |
| India (English) | 4.5 |
| **Average Bot Turns per Conversation** (5154 Total Turns, 368093 Total Words) | |
| U.S. (English) | 6.4 |
| U.S. (Spanish) | 5.7 |
| India (English) | 5.6 |
| **Issues Covered** | 8 |
| **Political Parties** | 4 |

Table 1: Dataset summary

### 3.3 Feature Extraction

Our choice of the conversation features are motivated by prior research on *computers as social actors* (CASA), which demonstrates that people routinely apply social heuristics, norms, and expectations to computational systems, even when they are aware that those systems are artificial (Nass et al., 1994). When technologies exhibit conversational cues such as turn-taking, responsiveness, or explanation, users are more likely to treat them as social interlocutors rather than neutral tools. In political contexts, this social framing is especially consequential, as attitudes, confidence, and knowledge are often shaped through interaction, perceived responsiveness, and epistemic trust rather than information exposure alone.

Therefore, we characterized conversational dynamics through a set of linguistic, explanatory, and engagement-related features from both chatbot and participant utterances. Features not directly collected via questionnaires were computed using established lexical resources (e.g., LIWC) and automated text-processing pipelines. For analysis, features were grouped into theoretically motivated blocks corresponding to controls, LLM explanatory behavior, and human engagement. Full operational definitions are provided in Appendix A.

- **Demographic and user-level control**: We included age, gender, income, country, conversation language, political ideology, and political efficacy, a measure of participants' perceived ability to understand and engage with political issues. These variables serve as controls to account for individual differences that may shape learning and confidence outcomes.
- **LLM explanatory features**: LLM explanatory behavior was operationalized through a set of linguistic and discourse-level features capturing explanation quantity, complexity, and informational structure. These included measures of syntactic complexity (e.g., average sentence length, conjunction and preposition use), readability (Flesch–Kincaid grade level), and information density (e.g., entity density, idea density). We also captured interactional characteristics such as conversation length, the proportion of explanatory content, and the use of informal language.

- **User Engagement features**: Participants' engagement during the conversation was assessed using LIWC-based linguistic markers associated with cognitive and epistemic processing. These included indicators of insight and reasoning (e.g., INSIGHT, CAUSE), tentativeness and uncertainty (TENTAT, DISCREP), and expressions of certainty or differentiation (CERTAIN, DIFFER). These features capture how actively participants processed information, articulated uncertainty, and reflected on the content during interaction with the LLM.

To obtain a more stable representation of engagement, we conducted an exploratory factor analysis on these variables. The analysis supported a single-factor solution, indicating that these linguistic markers load onto a common underlying construct. We therefore aggregated them into a composite measure, which we refer to as Cognitive Engagement, and used this variable in subsequent mediation and moderation analyses.

### 3.4 Statistical Analysis

Because our research questions concern both explanatory mechanisms and conditional effectiveness, we employ a combined mediation–moderation strategy rather than relying on main-effect models alone. These allowed us to (a) identify processes through which LLM explanatory features affect confidence and knowledge outcomes, and (b) test conditions under which these effects are amplified or attenuated by participant characteristics and engagement, providing a nuanced account of how LLM explanations function in human–LLM dialogue.

#### 3.4.1 Mediation Analyses

We conducted mediation analyses to test whether the effects of LLM explanatory quality on participant outcomes operated indirectly through user engagement processes. Based on theoretical relevance and prior empirical findings (), LLM explanatory ratio was selected as the focal LLM predictor across all mediation models as it captures the proportion of LLM responses devoted to explanatory content and has been shown to reflect explanation richness rather than surface verbosity. Mediator choice was guided by both theoretical alignment and empirical support:

- For confidence change, user insights was selected as the primary mediator. Insight reflects participants' reflection and sense-making during interaction, which are central to confidence formation. Confidence judgments are theorize to arise from internal appraisal of understanding rather than interaction alone.
- For knowledge gain, cognitive engagement was selected as the mediator. Cognitive engagement captures elaboration and effortful processing—mechanisms widely recognized as prerequisites for learning.

Mediation models were estimated using structural equation modeling with maximum likelihood estimation. Indirect effects were assessed using bootstrapped confidence intervals (5,000 resamples) to account for non-normality. Participant-level clustering was incorporated to adjust for repeated observations.

#### 3.4.2 Moderation Analyses

To examine whether the effects of LLM features on outcomes varied as a function of participant characteristics, we conducted moderation analyses using multilevel linear mixed-effects models with random intercepts for participants.

User Insights and Political Efficacy were moderators. Moderator–predictor interactions were selected based on theoretical coupling between conversational signals and outcome domains:

- For confidence change, moderators were interacted with help-seeking dialogue acts, reflecting moments of epistemic uncertainty. Confidence is expected to be particularly sensitive to how uncertainty is experienced and resolved.
- For knowledge gain, moderators were interacted with LLM conversation length, a proxy for exposure and opportunity for elaboration. Learning benefits from explanation are theorized to depend on sustained engagement rather than isolated clarification events.

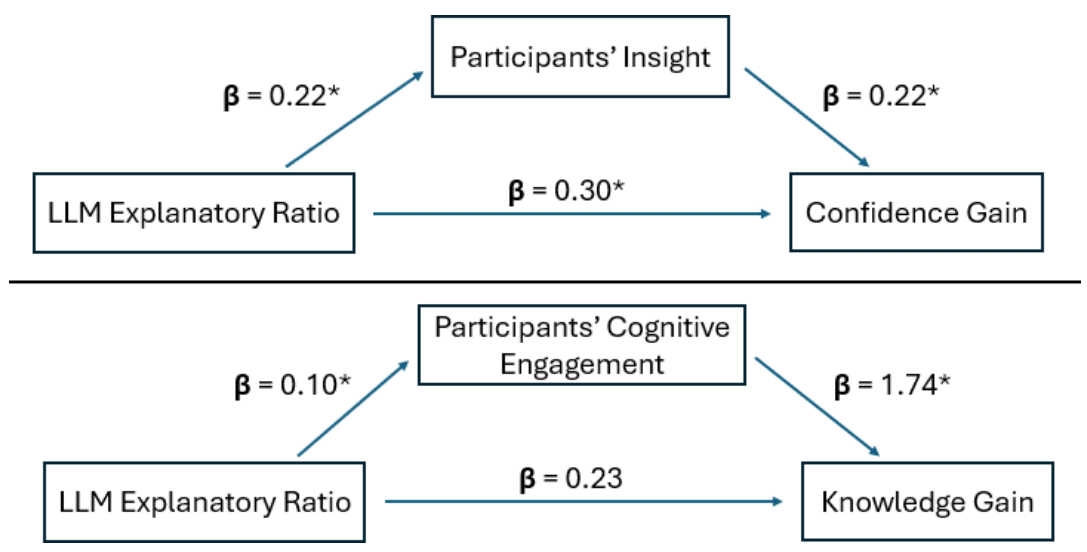

Figure 1: Mediation analyses results and path coefficients

## 4 Results

### 4.1 Mediation analysis

For each outcome, we compared a partial mediation model, which included both direct and indirect paths from explanatory ratio to the outcome, against a full mediation model, which included only the indirect path through the mediator.

#### 4.1.1 Confidence Gain

A likelihood ratio test comparing the full mediation model to the partial mediation model was significant (($\chi^2_{diff}$ = 18.35, $df = 1$, $p < .05$), indicating that the partial mediation model provided a superior fit. This result suggests that participants' insight accounts for part—but not all—of the association between LLM explanatory ratio and confidence change, consistent with a partial mediation pattern. Results from the partial mediation model (Figure 1) showed a significant total effect of explanatory ratio on confidence ($z = 4.58$, $p < .05$), indicating that higher proportions of explanatory content from the LLM were associated with greater increases in participants' confidence. Importantly, the indirect effect via participants' insight was also significant ($z = 2.06$, $p < .05$), suggesting that one pathway through which explanatory behavior supports confidence is by fostering reflective understanding during the interaction. However, the persistence of a direct effect implies that confidence gains may also arise from exposure to explanations, beyond what is captured by participants' expressed insight.

#### 4.1.2 Knowledge Gain

The likelihood ratio test comparing the partial and full mediation models was not significant ($\chi^2_{diff}$ = 0.41, $df = 1$, $p = .52$), indicating that the more parsimonious full mediation model fit the data equally well. This finding supports a full mediation interpretation, in which the effect of explanatory ratio on knowledge operates entirely through participants' cognitive engagement.

Consistent with this interpretation, the indirect effect via cognitive engagement was significant ($z$ = -2.25, $p < .05$), whereas the total effect of explanatory ratio on knowledge gain was not ($z$ = -1.64, $p$ = .10). Inspection of the component paths further revealed that explanatory ratio significantly predicted cognitive engagement, but did not directly predict knowledge gain (Figure 1). This indicate that explanatory richness contributes to learning primarily by shaping how cognitively engaged participants are during the conversation, rather than exerting a direct influence on knowledge outcomes.

### 4.2 Multi-level Analysis of Knowledge Gain and Confidence

Building on the mediation analyses, we next examined the conditions under which these mechanisms operate most effectively. Specifically, we tested whether the effects of LLM explanations on the outcomes varied as a function of (a) interactional engagement (e.g., insight and confusion/help-seeking dialogue acts) and (b) individual differences in users' orientations, such as political efficacy. To this end, we estimated a series of nested multilevel models, sequentially adding (1) demographic controls, (2) LLM explanatory features, and (3) theoretically motivated interaction terms.

#### 4.2.1 Confidence Gain

Adding the LLM explanatory block—comprising features such as average sentence length, explanatory ratio, readability, and other linguistic and discourse-level characteristics—significantly improved model fit relative to the demographic control model ($\chi^2_{diff}$ = 22.10, $df = 11$, $p < .05$; see Figure 2). This result indicates that variation in the linguistic and explanatory properties of the LLM contributes meaningfully to changes in participants' confidence beyond demographic factors alone.

We then examined whether confidence gains depended on users' engagement and efficacy. Including the interaction between users' insight and confusion/help-seeking acts improved model fit ($\chi^2_{diff}$ = 6.85, $df = 1$, $p < .05$). Figure 3 shows that confusion/help-seeking positively predicts confidence gain for high-insight users, but negatively for low-insight users. This suggests that confidence benefits from LLM explanations when users actively reflect on their uncertainty rather than passively seeking help.

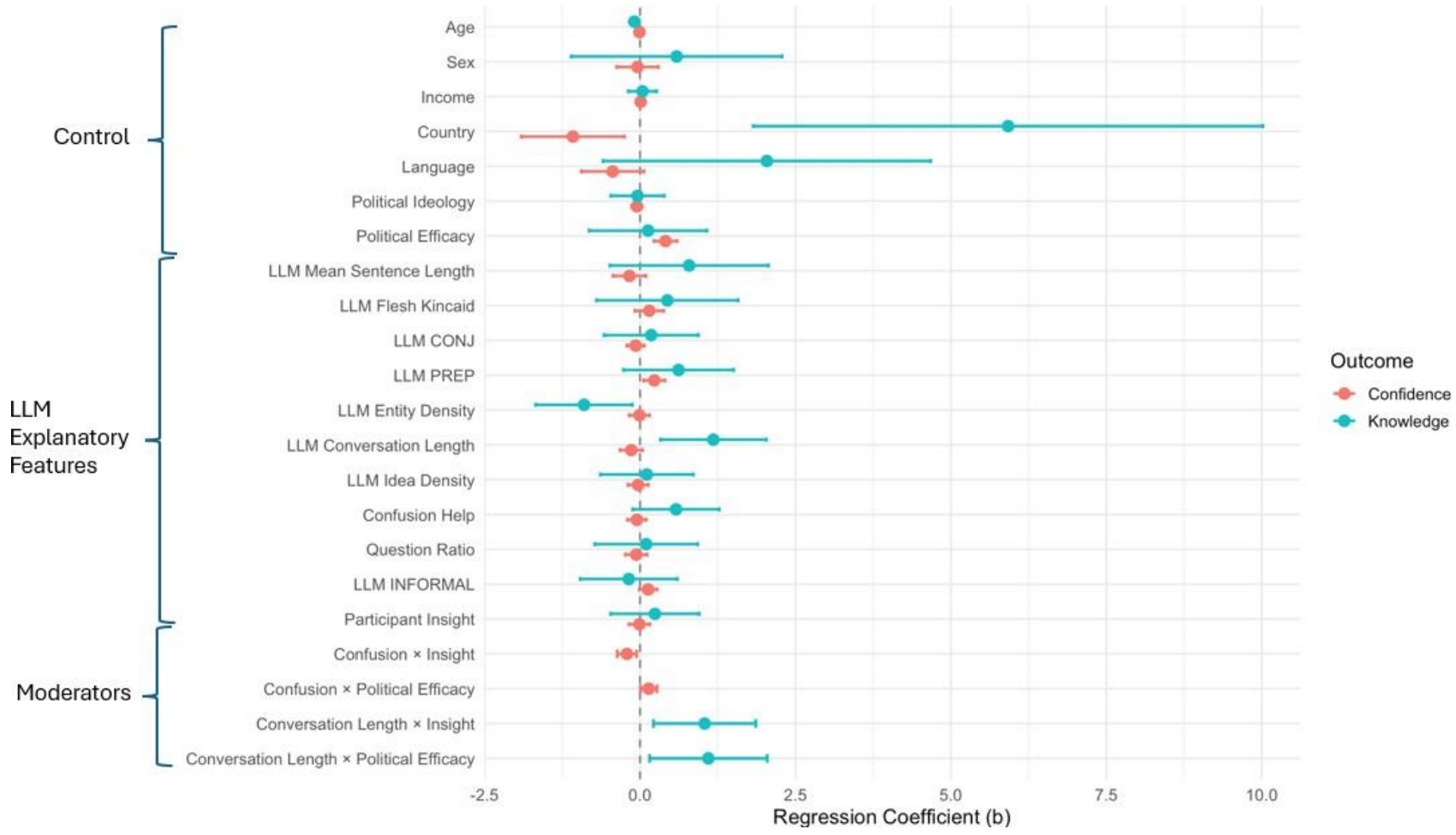

Figure 2: Coefficients from multilevel models predicting changes in confidence and knowledge, and moderator analyses results.

A similar interaction emerged for political efficacy ($x^2_{diff}$ = 4.44, $df$ = 1, $p$ < .05), where confusion/help-seeking increased confidence mainly for low-efficacy users, with no effect for high-efficacy users. Together, these findings indicate that confidence development depends on the alignment between engaged conversational behavior and users' perceived ability to interpret complex information.

### 4.2.2 Knowledge Gain

As with confidence, adding the LLM explanatory block significantly improved model fit relative to the control-only model ($x^2_{diff}$ = 21.41, $df$ = 11, $p$ < .05), indicating that explanatory and linguistic features of the LLM are systematically related to learning outcomes.

The interaction between insight and conversation length significantly improved model fit ($x^2_{diff}$ = 6.39, $df$ = 1, $p$ < .05). Simple slopes analyses indicated that conversation length was positively associated with knowledge gain at high levels of insight, whereas the association was weak and non-significant at low levels of insight (Figure 3). This pattern suggests that reflective capacity may enable users to leverage longer conversational exchanges more effectively, translating extended interaction into cumulative knowledge gains.

The interaction between political efficacy and conversation length was also significant ($x^2_{diff}$ = 5.80, $df$ = 1, $p$ < .05). As shown in Figure 3, longer conversations were associated with greater knowledge gains among high-efficacy users, whereas conversation length was not significantly related to knowledge gain for users with low political efficacy. Users with higher political efficacy may engage more actively and purposefully in longer conversations, allowing them to convert sustained interaction into learning gains.

Together, these interactions indicate that extended conversational engagement is most beneficial for users with greater reflective and motivational resources, while offering limited gains for users lower on these dimensions.

These moderation results complement prior mediation analyses, indicating that LLM explanations enhance learning primarily by sustaining engaged, reflective interaction, rather than providing uniform benefits across all users. Altogether, the findings highlight that the educational impact of LLMs emerges from the interplay between explanatory design, conversational engagement, and individual cognitive capacities.

## 5 Discussion

Much of the literature on LLMs in education and political communication implicitly assumes that conversational systems exert broadly similar effects across users, or that richer explanations should reliably improve learning. Our findings challenge this assumption. Explanations matter, but primarily insofar as they structure engagement. Confidence can be shaped through exposure to explanations that signal coherence or competence.

LLMs' social responses are not merely attitudinal or affective, but have structured consequences for learning. Importantly, however, these consequences are not uniform. Rather than functioning as generic social actors, LLMs appear to operate as *conditional social partners*: their influence on confidence and knowledge depends on how users engage with them and on users' capacities to interpret, reflect on, and sustain interaction. Knowledge, by contrast, depends more tightly on whether users actively elaborate on and integrate information during dialogue. This divergence highlights why confidence and knowledge should not be treated interchangeably when evaluating conversational AI.

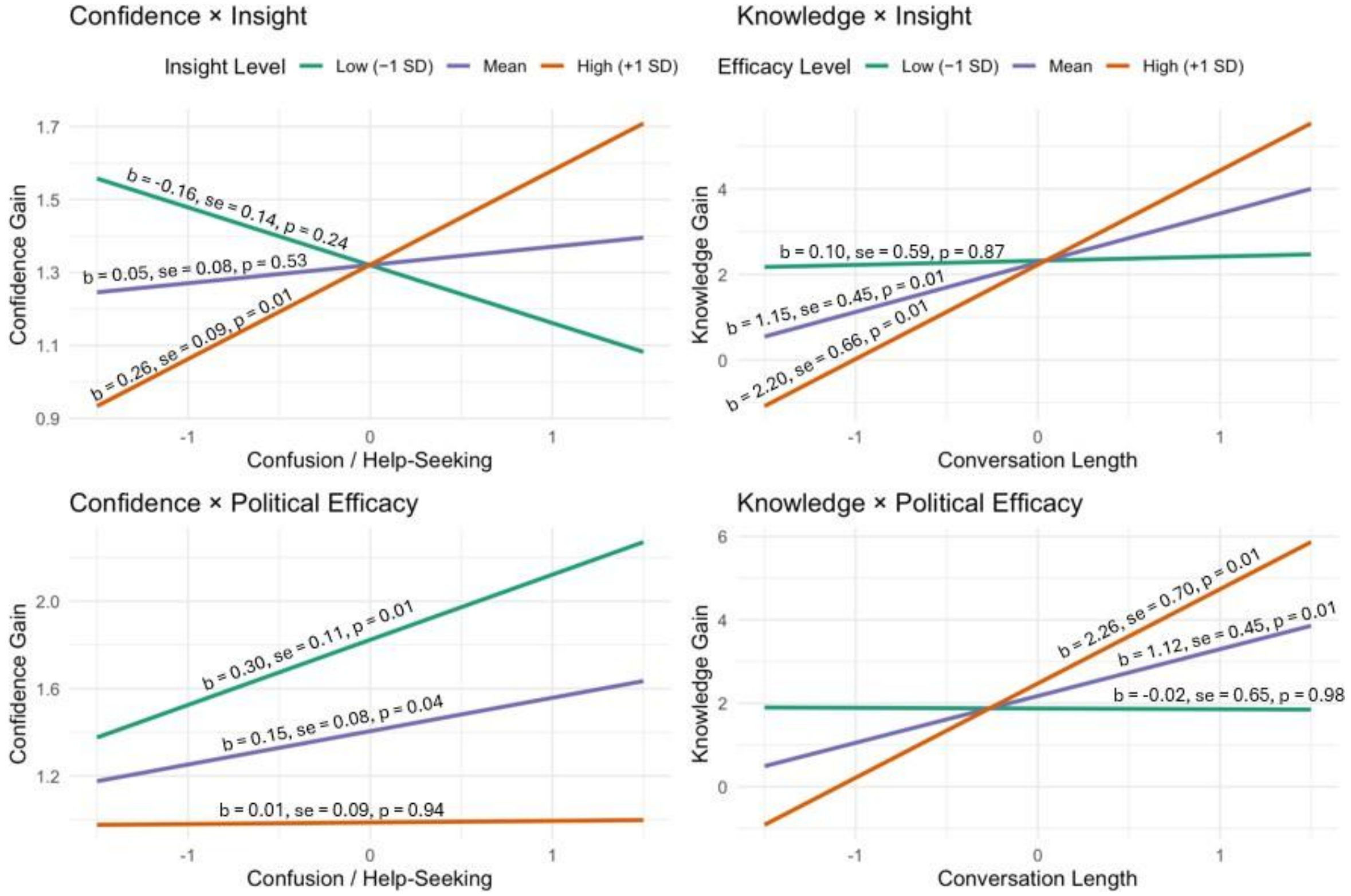

Figure 3: Simple slopes for significant moderation effects. Panels 1–2 show the interaction between confusion/help-seeking and participant characteristics predicting confidence gain. Panels 3–4 show the interaction between LLM conversation length and participant characteristics predicting knowledge gain. Lines represent predicted outcomes at low (−1 SD), mean, and high (+1 SD) levels of the moderator.

The moderation results further underscore the need to move beyond average treatment effects. Interactional cues such as uncertainty, help-seeking, and conversational depth do not have intrinsic valence; their effects depend on users' reflective capacity and political efficacy. From an interactional standpoint, extended dialogue is productive only when users are able to leverage it. These patterns are difficult to reconcile with output-centric or persuasion-focused evaluations, but follow naturally from theories that treat learning as a collaborative and effortful process.

## 6 Conclusion

This study contributes to a growing body of work on conversational AI with a process-oriented account of learning in political dialogue to show that learning emerges from interactional alignment: explanations support confidence and knowledge only when they elicit forms of engagement that users are able to sustain and interpret productively.

More broadly, our findings extend the computers-as-social-actors perspective by demonstrating that conversational LLMs can function as *differentiated social partners*. Their effects depend on users' reflective and motivational resources, as well as on how uncertainty, explanation, and conversational depth are coordinated during interaction. Consequently, they can increase users' confidence without reliably improving knowledge, and extended dialogue may benefit some users while offering limited gains for others. For instance, future conversational agents may want to scaffold reflection where needed and avoiding unnecessary conversational complexity when it is unlikely to support learning.

## Limitations

Our experimental design captures short-term changes in users' knowledge and confidence but does not address longer-term effects on belief consolidation, political behavior, or sustained resilience to misinformation. Although the study spans two national contexts and multiple languages, the findings may not generalize to other political systems, media environments, or demographic groups. In addition, the Spanish-language condition relied on a more limited pool of available news sources, which may have constrained the range of perspectives presented and influenced conversational dynamics.

While the use of pre–post outcome measures strengthens temporal ordering, the mediation and moderation analyses remain associative rather than causal. Explanatory features and engagement processes are not experimentally manipulated and may co-evolve with unobserved user characteristics or interactional factors. As a result, the identified pathways and conditional effects should be interpreted as statistically grounded, process-consistent associations rather than definitive causal mechanisms.

## Ethical considerations

The study protocol was reviewed and approved by the Institutional Review Board (IRB) of our university. All participants provided informed consent prior to participation and were informed of the study's purpose, their right to withdraw at any time, and the handling of their data. No potentially identifying information was collected. All participants were debriefed about the true nature of the treatment at the end of the survey, following the protocol approved by the IRB.

Because chatbot interactions could involve sensitive political beliefs and reasoning expressed in free-form text, ethical and privacy considerations were integrated directly into the system's technical design. Model inference was conducted via Groq's inference-only API, which guarantees that user prompts and model outputs are not stored, logged, or reused for model training by the provider, and are retained only transiently to fulfill each request. This ensured that no participant data entered external training or analytics pipelines.

All persistent logging—including conversation transcripts and experimental metadata—was handled exclusively by a custom backend and stored on institutionally approved, university-owned Amazon Web Services infrastructure, where survey responses were stored on a separate server and linked via anonymous identifiers to the chat logs. The system was designed to minimize data exposure by restricting access to interaction logs to the research team and avoiding any third-party data storage or reuse. These design choices ensured that privacy protection was enforced at the architectural level, independently of downstream data handling procedures described in the following section.

## Acknowledgments

This research was supported by the Ministry of Education, Singapore, through its MOE AcRF Tier 3 Grant (MOE- MOET32022-0001) and the Tier 1 programme (WBS A-8000231-01-00).

# A Feature Definitions, Operationalization, and Sources

This appendix provides detailed descriptions of all variables included in Table 1. Variables are grouped according to their conceptual role in the analysis, following the organization used in the Feature Extraction section: (1) demographic and user-level controls, (2) LLM explanatory features, and (3) human engagement and interactional features. For each variable, we describe its definition, how it was computed, and the tools or resources used.

## A.1 Demographic and User-Level Variables

**Age:** Participant self-reported age in years, collected via pre-conversation questionnaire.

**Sex:** Self-reported participant sex, collected via questionnaire.

**Income:** Self-reported household income bracket, collected via questionnaire and treated as an ordinal variable.
**Country:** Binary indicator of participant national context (United States vs. India), determined by study assignment.
**Language:** Conversation language (English vs. Spanish). Language was fixed by experimental condition and used as a control to account for linguistic variation.
**Political Ideology:** Self-reported political ideology measured on a Likert-style scale ranging from liberal (1) to conservative (6) (or left–right equivalent depending on country context).
**Political Efficacy:** A composite self-report measure capturing participants' perceived ability to understand and engage with political issues. Political efficacy is widely used in political psychology as a predictor of information processing and confidence formation.

### A.2 LLM Explanatory Features

These features characterize the linguistic complexity, informational structure, and explanatory style of the LLM's responses. Unless otherwise noted, features were computed at the conversation level by aggregating across all LLM turns.
**LLM Mean Sentence Length:** Average number of words per sentence in LLM responses. This metric captures syntactic complexity and verbosity.
**LLM Flesch–Kincaid Readability:** Flesch–Kincaid grade level score computed using the *textstat* library. Higher values indicate more linguistically complex explanations requiring greater reading proficiency.
**LLM CONJ (Conjunction Use):** Proportion of conjunctions in LLM responses, derived from LIWC. Conjunctions are often associated with causal and elaborative explanation structures.
**LLM PREP (Preposition Use):** Proportion of prepositions in LLM responses, derived from LIWC. Preposition usage reflects syntactic embedding and informational density.
**LLM Entity Density:** Ratio of unique named entities (persons, organizations, locations) to total word count in LLM responses. Entities were extracted using standard named entity recognition pipelines (spaCy). This measure captures informational specificity.
**LLM Conversation Length:** Total number of conversational turns exchanged between the participant and the LLM. Conversation length serves as a proxy for exposure and opportunity for elaboration.

**LLM Idea Density:** A measure of propositional content per word, capturing how many distinct ideas are expressed relative to message length. Higher values indicate denser informational content. Extracted using the **ideadensity** python library.

**LLM Informality:** Degree of informal language use in LLM responses, measured using LIWC categories associated with informal, colloquial, or conversational markers.

**Question Ratio:** Proportion of turns containing questions. This metric captures the extent to which the LLM adopts an inquiry-driven or dialogic explanatory style.

### A.3 Human Engagement and Interactional Features

These features capture how participants engaged cognitively and interactionally during the conversation. All measures are derived from participant utterances unless otherwise noted.

**Participant Insight (Main Effect)** LIWC-based measure capturing expressions of insight, reflection, and understanding (e.g., words such as *realize*, *understand*, *think*). Insight is commonly associated with sense-making processes and confidence formation.

**Cognitive Engagement (Composite):** A composite measure derived from exploratory factor analysis of multiple LIWC cognitive-process categories, including INSIGHT, CAUSE, DISCREP, TENTAT, CERTAIN, and DIFFER. This factor reflects elaborative processing, reasoning, and epistemic effort during dialogue.

**Confusion / Help-Seeking:** Frequency of participant dialogue acts indicating confusion, uncertainty, or explicit requests for clarification. These acts mark moments of epistemic breakdown and opportunities for explanation.

## B Chatbot Prompt Templates

**Prompt Instructions (Bot Only – Shared Across Contexts):**
You are an AI assistant designed to support political information-seeking through dialogue. You will be provided with a list of news articles and a list of key concepts in question format.

- Use **only the information contained in the provided articles**. Do not introduce outside facts or deviate from the given sources.
- Address **one key concept at a time**. Explain the concept and why it matters in the given political context.
- Take the learning process slow. Use simple language, short sentences, and clear explanations.
- Invite the user to reflect by asking **two open-ended questions** about key concepts you want to discuss next. Encourage longer, thoughtful responses; avoid yes/no questions when possible.
- If the user asks about sources, provide **only** the list of articles supplied. Never reveal the list of key concepts.
- After **5 turns**, summarize the discussion and ask whether the user would like to close the conversation or continue.
- Example: “So did you know about ?” - If the user’s response is short or vague, nudge them with questions like: “Do you think...?” -
- Base your response on the following context: - List of articles: news[’articles’] - Key concepts: news[’summary’] - State: news[’state’] - Issue: news[’issue’] - Country: news[’country’]

- Generate a response that fits this format while addressing the user’s most recent message: “user$_{m}$*essage*”

**Context: USA (English)**

**Opening Comment (Bot):**
Hello, I am a news summarisation assistant. I will be answering the query about policies and achievements of PARTY on ISSUE in STATE in the election of 2024. Is there anything particular that you want to start with?

**Context: USA (Spanish)**
**Special Prompt Instruction:**
You will converse only in Spanish.
**Opening Comment (Bot):**
Hola, soy asistente de resumen de noticias. Responderé la consulta sobre las políticas y los logros del PARTY sobre el ISSUE en el USA en las elecciones de 2024. ¿Hay algo en particular con lo que quiera comenzar?

**Context: India (English)**
**Special prompt instruction**:
The Indian elections have already happened, so share information or ask questions in the past tense, e.g. “What do you think was an important issue...?”

**Opening Comment (Bot):**
Hello, I am a news summarisation assistant. I will be answering the query about policies and achievements of PARTY on ISSUE in STATE in the election of 2024. Is there anything particular that you want to start with?

## C Exemplar Knowledge Questions

The style and types of questions asked were based on prior characterizations of information-seeking questions on the ComQA dataset (Abujabal et al., 2019).

### C.1 USA English questions about Democratic Party on Air Quality

**Simple Questions**:

**Q1.** What topic is predicted to be a significant factor for voters in the 2024 election?

A) Education reform
B) Climate change
C) Healthcare policies
D) Tax reform

**Answer:** B) Climate change

**Q2.** What has driven younger voters to prioritize certain issues in recent elections?

A) Economic incentives
B) Technological advancements
C) Environmental concerns
D) Social media influence

**Answer:** C) Environmental concerns

**Compositional Questions**

**Q1.** Which environmental initiatives are highlighted in the 2024 midterm elections?

A) Renewable energy and EV incentives
B) Water conservation and ocean preservation
C) Industrial tax breaks and economic incentives
D) Urban development and housing reforms

**Answer:** A) Renewable energy and EV incentives

**Q2.** What aspect of Biden’s environmental policy has been deemed contradictory?

A) His focus on both climate action and fossil fuel independence
B) His support for reducing pollution alongside increased plastic production
C) His push for renewable energy while increasing taxes on solar products
D) His dedication to climate change policies without support for electric vehicles

**Answer:** A) His focus on both climate action and fossil fuel independence

**Temporal Questions**

**Q1.** When did four major environmental groups endorse President Biden's reelection?

A) Before his State of the Union address
B) During the League of Conservation Voters dinner
C) Following the 2020 election
D) During the 2022 midterm elections

**Answer:** B) During the League of Conservation Voters dinner

**Q2.** In which past elections did climate change concerns reportedly influence voter choices?

A) 2016 and 2020
B) 2008 and 2012
C) 2012 and 2016
D) 2000 and 2004

**Answer:** A) 2016 and 2020

**Comparison Questions**

**Q1.** How do the environmental concerns of younger voters compare to those of older voters in recent elections?

A) Younger voters prioritize climate change, while older voters focus more on inflation.
B) Older voters are more concerned about climate change than younger voters.
C) Both age groups equally prioritize renewable energy initiatives.
D) Younger voters are less concerned about climate change than older voters.

**Answer:** A) Younger voters prioritize climate change, while older voters focus more on inflation.

**Q2.** Compared to 2020, what stance are Democratic candidates likely to maintain on climate-related issues in 2024?

A) Less aggressive due to economic concerns
B) Similar, with emphasis on clean energy jobs
C) Stricter focus on fossil fuel independence
D) Reduced focus on renewable energy projects

**Answer:** B) Similar, with emphasis on clean energy jobs

**Answer Tuple Questions**

**Q1.** Identify the two main topics highlighted as climate-related factors in the 2024 elections. (Choose all that apply)

A) Clean energy jobs
B) Corporate tax cuts
C) Fossil fuel independence
D) Transportation funding

**Answer:** A) Clean energy jobs, C) Fossil fuel independence

**Q2.** Which organizations have shown significant ties to Biden's environmental policies? (Choose all that apply)

A) League of Conservation Voters
B) Green Peace
C) Environmental Protection Agency (EPA) allies
D) U.S. Chamber of Commerce

**Answer:** A) League of Conservation Voters, C) Environmental Protection Agency (EPA) allies